\documentclass[letterpaper, 10 pt, conference]{ieeeconf}  

\IEEEoverridecommandlockouts                              

\usepackage{graphicx}
\usepackage{amsmath}
\usepackage{amssymb}
\usepackage{booktabs}

\usepackage{color}
\usepackage{hyperref}
\title{\LARGE \bf
Learning Generalizable 3D Manipulation With 10 Demonstrations
}

\author{Yu Ren, Yang~Cong,~\IEEEmembership{Senior~Member,~IEEE,}, Ronghan Chen, and Jiahao Long 
\thanks{Yu Ren and Ronghan Chen are with the State Key Laboratory of Robotics, Shenyang Institute of Automation, Chinese Academy of Sciences, Shenyang 110016, China and also with the University of Chinese Academy of Sciences, Beijing 100049, China (e-mail: renyu0414@gmail.com, chenronghan@sia.cn).}%
\thanks{Yang Cong and Jiahao Long are with the College of Automation Science and Engineering, South China University of Technology, Guangzhou 510640, China (email: congyang81@gmail.com, 202420115946@mail.scut.edu.cn). Corresponding author: Prof. Yang Cong.}%
\thanks{}%
}

\begin{document}

\maketitle
\thispagestyle{empty}
\pagestyle{empty}

\begin{abstract}

Learning robust and generalizable manipulation skills from demonstrations remains a key challenge in robotics, with broad applications in industrial automation and service robotics. While recent imitation learning methods have achieved impressive results, they often require large amounts of demonstration data and struggle to generalize across different spatial variants. In this work, we present a novel framework that learns manipulation skills from as few as 10 demonstrations, yet still generalizes to spatial variants such as different initial object positions and camera viewpoints. Our framework consists of two key modules: Semantic Guided Perception (SGP), which constructs task-focused, spatially aware 3D point cloud representations from RGB-D inputs; and Spatial Generalized Decision (SGD), an efficient diffusion-based decision-making module that generates actions via denoising. To effectively learn generalization ability from limited data, we introduce a critical spatially equivariant training strategy that captures the spatial knowledge embedded in expert demonstrations. We validate our framework through extensive experiments on both simulation benchmarks and real-world robotic systems. Our method demonstrates a 60–70\% improvement in success rates over state-of-the-art approaches on a series of challenging tasks, even with substantial variations in object poses and camera viewpoints. This work shows significant potential for advancing efficient, generalizable manipulation skill learning in real-world applications.\footnote{ https://github.com/renyu2016/Generalized-3D-Manipulation}

\end{abstract}

\section{INTRODUCTION}

Learning robust and generalizable manipulation skills from demonstrations\cite{wen2022catgrasp, WenLBS22, ren2023autonomous, zhi2024instructing} is a longstanding goal in robotics research, with broad applications in various aspects of human life, such as industrial robot assembly \cite{morgan2021vision, zhu2024multi} and service robots performing housework \cite{wen2022you, ren2023autonomous}. Recently, several imitation learning methods\cite{papagiannismiles, jiamail, chi2023diffusionpolicy} have shown remarkable performance in learning manipulation skills. However, these methods still suffer from the need for large amounts of demonstration data and exhibit limited generalization ability.

Recent approaches combining 3D representations with Diffusion Policy \cite{Ze2024DP3, chi2023diffusionpolicy, lu2024manicm} have shown potential performance in learning manipulation tasks from limited demonstrations, generalizing to different visual appearances, instance geometries, and camera viewpoints. However, these methods tend to overfit specific training trajectories rather than capturing the spatial relationships\cite{Lee24, LuoLWDL24, abs-2406-01029, shridhar2022cliport, LuGWBZR23} needed for generalization. This limitation leads to poor performance in tasks with varying initial object and target positions, or when the starting pose of a manipulated object significantly deviates from the scenes contained in training data, \emph{i.e.,} these methods lack the 3D generalization ability.

\begin{figure}[t]
	\centering
	\includegraphics[trim = 43mm 130mm 39mm 60mm, clip, width=0.5\textwidth, height=210pt]{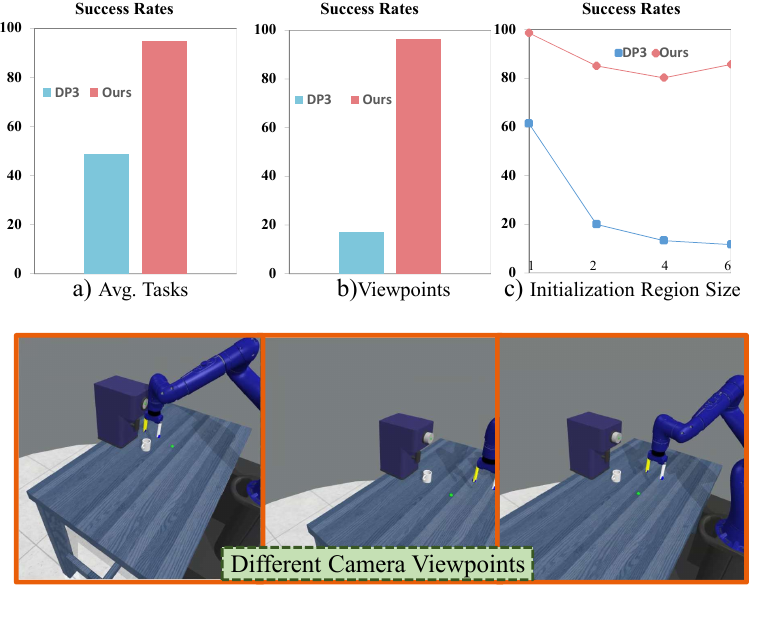}
	\vspace{-35pt}
	\caption{In a), we report the average success rates of the two methods on a series of challenging tasks. In b), we show the average success rates of the two methods under different camera viewpoint settings. In c), we progressively expand the random initialization region where the manipulated object is located at the start of the tasks. Compared to the state-of-the-art 3D manipulation learning method DP3, our framework demonstrates significant improvements in success rates and generalization ability.}
	\vspace{-20pt}
	\label{fig:ref}
\end{figure}

 To efficiently learn manipulation skills from a few demonstrations while ensuring 3D generalization, we propose a framework that leverages the spatial information in demonstrations to achieve 3D-generalized manipulation skill learning. Our framework consists of two key components:  \textbf{S}emantic \textbf{G}uided \textbf{P}erception (\textbf{SGP}) and \textbf{S}patial \textbf{G}eneralized \textbf{D}ecesion (\textbf{SGD}). SGP constructs a 3D representation from RGB-D image pairs that is both task-focused and spatially aware. SGD is an efficient decision-making module based on diffusion policy\cite{kapelyukh2023dall, chen2023playfusion, mishra2023generative, ScheiklSHFNLM24, LiBSR24}. To enable SGD to generalize across different 3D spatial variations, we introduce a spatially equivariant training strategy that fully explores the spatial knowledge embedded in expert trajectories.
 
 
To evaluate our framework, we conducted extensive experiments on both simulation benchmarks and real-world hardware system. Using the same task settings as current state-of-the-art methods, our approach achieved a \textbf{60-70\%} improvement in success rates on a series of challenging simulation tasks with only 10 demonstrations. Additionally, we designed comprehensive experiments to demonstrate the strong \textbf{generalization ability} of our framework in handling spatial variations and viewpoint changes.
Our contributions are summarized as follows:
\begin{itemize}
    \item We propose a framework that learns 3D generalized manipulation skills with only 10 demonstrations. Our framework generalizes to varying object initial poses and camera viewpoints.
     \item We develop an easy to use yet highly effective training strategy for manipulation policies, enabling the exploration of spatial knowledge embedded in demonstration trajectories. This training strategy is easily integrated with any frameworks that use point cloud as input for decision policy learning.

    \item We validate our framework through extensive simulation and real-world experiments, achieving over a 60\% improvement on a series of challenging tasks compared to state-of-the-art methods, demonstrating its strong effectiveness.
\end{itemize}

\section{Related Work}

\subsection{Visual Imitation Learning}

Recently, several imitation learning methods \cite{fu2024mobile, chi2023diffusionpolicy, chi2024universal, ha2023scaling, xu2023xskill} have demonstrated remarkable performance in learning manipulation skills. These methods typically take RGB-D images as input and utilize a decision policy $\pi$ to map the current or recent observations to appropriate actions, allowing the robot to successfully complete tasks \cite{chi2023diffusionpolicy}. However, learning such policies usually requires a large amount of demonstration data, which can be expensive and time-consuming to collect \cite{Ze2024DP3}. Moreover, 2D image-based methods are sensitive to variations in visual appearance, lighting, and observation angles. If the testing environment differs from the training environment, performance often degrades significantly.
To address these issues, recent methods have adopted 3D representations, such as point cloud \cite{Ze2024DP3, xian2023chaineddiffuser, gervet2023act3d, zhulearning, ma2024hierarchical,zhu2024vision}, as input for action prediction and generation. Notably, DP3 \cite{Ze2024DP3} combines point cloud representation with diffusion model to learn manipulation skills from just 10 demonstrations, achieving generalization across variations in visual appearance, object geometry, and viewpoint changes. While effective, the model has been observed to overfit the seen trajectories when trained with limited data. This leads to failures when the manipulated objects have different initial or target poses that require new trajectories. Therefore, learning generalizable manipulation from a few demonstrations remains an open challenge.


\subsection{Diffusion Models in Robotics}

Diffusion models, a class of generative models \cite{zhang2023adding, tang2023emergent, tumanyan2023plug,ruiz2023dreambooth}that progressively transform random noise into data samples, have achieved significant success in high-fidelity image generation. Due to their expressive capabilities, diffusion models have recently been applied to robotic manipulation, where they represent a robot’s visuomotor policy as a conditional denoising diffusion process \cite{Ze2024DP3, lu2024manicm, xian2023chaineddiffuser, ma2024hierarchical}. Compared to directly predicting actions from observations using deep learning networks, the denoising approach excels at handling multimodal action distributions, is well-suited for high-dimensional action spaces, and demonstrates impressive training stability \cite{chi2023diffusionpolicy}. However, while diffusion models can establish stable mappings between observations and behaviors, they lack the ability to fully capture the spatial relationships between objects involved in manipulation tasks. Addressing how diffusion models can learn and utilize spatial knowledge from trajectories with limited data remains a critical challenge.


\subsection{Semantic Foundation Model}

Recently, foundational models for video \cite{kirillov2023segment, cheng2024putting} and language \cite{radford2021learning} have achieved impressive advancements, offering significant benefits for robotics research. These models allow the use of pre-trained weights, eliminating the need for extensive data collection and scene-specific training, thereby greatly simplifying robot perception and scene understanding\cite{ravi2024sam, li2023mask}. For instance, SAM \cite{kirillov2023segment} enables robots to effectively separate target objects from backgrounds, while CLIP \cite{radford2021learning} connects natural language with images, allowing robots to generalize across different object appearances. Cutie \cite{cheng2024putting}, a video segmentation foundational model, helps robots maintain focus on manipulated objects while learning or executing tasks. The generalization ability of these models, combined with their lack of need for scene-specific training, enables robots to leverage their capabilities to achieve generalization across environments with minimal training data.

\begin{figure*}[t]
	\centering
	\includegraphics[trim = 20mm 160mm 20mm 50mm, clip, width=1\textwidth, height=240pt]{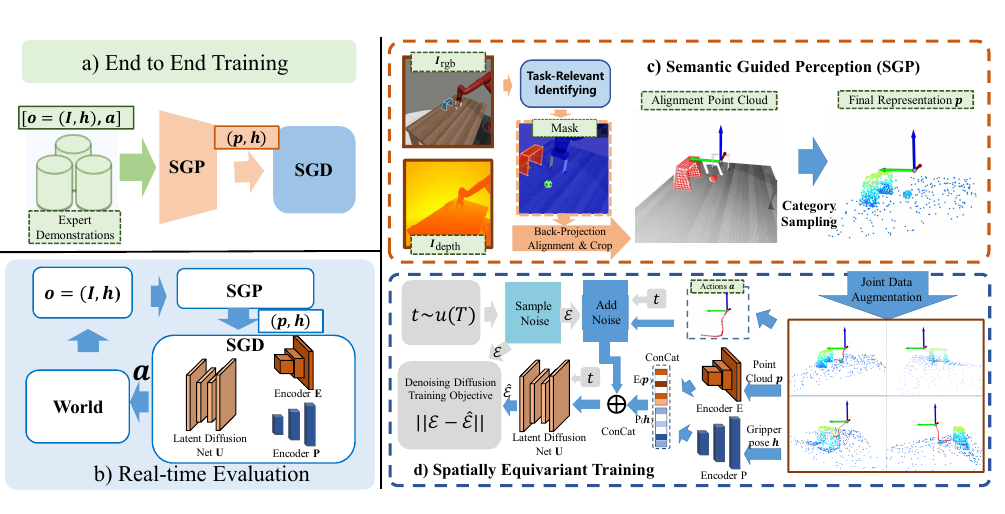}
	\vspace{-25pt}
	\caption{Overall workflow of our developed framework, which consists of two key modules: Semantic Guided Perception (SGP) and Spatial Generalized Decision (SGD). In (a) and (b), SGP takes the RGB-D images $\mathbf{I}$ and gripper pose $\mathbf{h}$ as inputs, constructing a 3D point cloud representation $\mathbf{p}$ that is both task-focused and spatially aware. The details of SGP are shown in (c). SGD is an efficient decision-making module based on diffusion policy. To enable our framework to generalize across different 3D spatial variations, we introduce a spatially equivariant training strategy that leverages the spatial knowledge embedded in expert trajectories, as illustrated in (d).}
	\vspace{-15pt}
    \label{framework_overall}
\end{figure*}

\vspace{-3pt}
\section{Method}
Given a small set of expert demonstrations $\mathcal{D} = \{\tau_i\}_{i=1}^{N} (N\leq10)$, where each trajectory $\tau_i$ consists of a sequence of observation-action pairs $(\mathbf{o}_1, \mathbf{a}_1,\ldots, \mathbf{o}_T, \mathbf{a}_T)$, we aim to learn a vision-based manipulation policy $\pi: \mathcal{O}\rightarrow\ \mathcal{A}$ that maps observations $\mathbf{o} \in \mathcal{O}$ to actions $\mathbf{a} \in \mathcal{A}$, allowing the robot not only to reproduce the demonstrated skills but also to generalize to observation states beyond the training data. In this paper, the observation $\mathbf{o}$ contains the RGBD image $\mathbf{I}$ captured by a camera with known camera parameters and the gripper pose $\mathbf{h}$ read from robot. The action $\mathbf{a}$ denotes the relative movement and the open or close state of the gripper at each step. 

\subsection{Overall framework}
To learn a manipulation policy $\pi$ from limited data while ensuring generalization to spatial variations, such as different initial object poses and changes in camera viewpoints, we developed the framework shown in Fig. \ref{framework_overall}. This framework consists of two key components: \textbf{S}emantic \textbf{G}uided \textbf{P}erception (SGP) and \textbf{S}patial \textbf{G}eneralized \textbf{D}ecision (SGD). As illustrated in Fig. ~\ref{framework_overall} (b), SGP takes RGB-D images $\mathbf{I}$ and gripper pose $\mathbf{h}$ from the world scene as inputs to construct a 3D point cloud representation $\mathbf{p}$ that is both task-focused and spatially aware. SGD, an efficient decision-making module based on diffusion policy\cite{chi2023diffusionpolicy}, uses $\mathbf{p}$ as input to generate the action $\mathbf{a}$ for the robot to execute through a denoising manner. This process can be formulated as:
\begin{equation}
    \mathbf{a} = SGD(\mathbf{p},\mathbf{h}) = SGD(SGP(\mathbf{I},\mathbf{h}),\mathbf{h})
\end{equation}

To enable SGD to generalize across various 3D spatial variations, we introduce a \textbf{Spatially Equivariant Training} strategy that leverages the spatial knowledge embedded in expert trajectories.


\subsection{Semantic Guided Perception (SGP)}
 SGP aims to construct representations from RGB-D image pairs that are both task-focused, emphasizing task-relevant objects while ignoring background information, and spatially aware, capturing the relative 3D spatial relationships between manipulation objects. To achieve this, given the observed RGB-D images $\mathbf{I}$ of current step, SGP constructs a task-relevant point cloud $\mathbf{p}$ through the following steps: 1) \textbf{Identifying} task-relevant objects and tracking them over time in the RGB images, 2) \textbf{Transforming} the point data into a predefined coordinate reference, and 3) Applying a \textbf{focused sampling} strategy for efficient feature embedding.

\textbf{Identifying task relevant objects.}
To identify relevant objects in an RGB image, we generate a mask $M$, where pixels of task relevant objects are assigned category IDs and background pixels are set to 0. This process starts with segmenting the first frame and tracking the mask over time using video segmentation methods. With the advancements in visual and natural language foundational models, there are multiple ways to obtain the first frame mask $M_0$. For example, SAM \cite{kirillov2023segment} can be used to segment objects interactively through clicks. Alternatively, it can be combined with open-vocabulary detection models to directly segment objects in the scene based on text descriptions of task-relevant objects. During evaluation, text descriptions can still be used to identify objects in new scenes, or visual correspondence methods can be applied to match the most similar mask between evaluation and training scenes, as demonstrated in \cite{zhulearning}.
\vspace{-1pt}

\textbf{Coordinates Reference Selecting.}
Given the RGB-D images $\mathbf{I}$ and task-relevant mask $M$, we construct a point cloud of the entire scene and align it to a predefined reference frame. Unlike previous works such as DP3 \cite{Ze2024DP3}, which use a fixed reference frame (e.g., the robot base), we select a dynamic frame, \emph{i.e.,} the gripper base at each timestep, as shown in Fig. \ref{fig:ref}. This approach significantly reduces learning complexity, as point clouds from different trajectories tend to converge to a similar state when using the gripper base as a reference, due to the consistent relative poses between the gripper and target objects at the end of the manipulation.
Next, we back-project the mask $M$ onto the point cloud, assigning each point a category label based on pixel-to-point correspondence.


\textbf{Category Points Sampling.}
Next, we crop and sample a sparse point cloud from the constructed scene for feature extraction. Instead of uniformly sampling across the entire point cloud, as done in works like DP3 \cite{Ze2024DP3, lu2024manicm}, we use the category ids from the previous step, sampling an equal number of points for background and all task-relevant objects. This approach makes the model pay more attention to task-relevant objects while reducing the influence of background variations. Unlike GROOT \cite{zhulearning}, which only uses the point cloud of target objects, our category-level sampling also work well when target objects are occluded, making the method more adaptable to different conditions.


\begin{figure}[t]
	\centering
	\includegraphics[trim = 27mm 90mm 40mm 70mm, clip, width=0.5\textwidth, height=220pt]{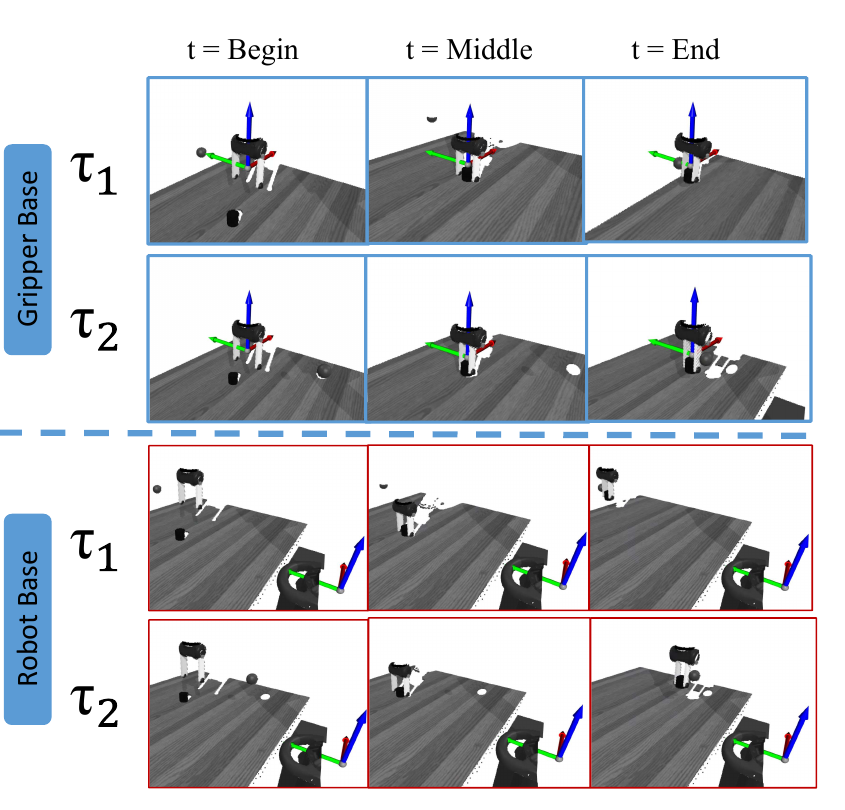}
	\vspace{-25pt}
	\caption{We compared two choices of coordinate reference. When using the robot base as the reference, the perceived points of the manipulated object vary significantly between trajectories due to differences in initial and target poses. In contrast, when using the gripper base as the reference, the points of the manipulated objects across different trajectories converge to the same state, as the relative pose between the gripper and the objects remains consistent.}
	\vspace{-15pt}
	\label{fig:ref}
\end{figure}



\subsection{Spatial Generalized Decision (SGD)}

SGD is a diffusion-based module \cite{chi2023diffusionpolicy, Ze2024DP3} that takes the point cloud $\mathbf{p}$ from SGP as input and generates actions $\mathbf{a}$ in a denoising process.  As shown in Fig. \ref{framework_overall} (b, d), after obtaining $\mathbf{p}$ and the gripper pose $\mathbf{h}$, SGD uses encoders $E$ and $P$ to embed these inputs into feature representations $\mathbf{E}(\mathbf{p})$ and $\mathbf{P}(\mathbf{h})$.

\textbf{During training phase} (Fig. \ref{framework_overall} (d)), random Gaussian noise is added to the ground truth actions $\mathbf{a}$. The Latent Diffusion Network $\mathbf{U}$ is trained to predict this noise, conditioned on the embedded features. The training objective is:
\begin{equation}
    \mathcal{L} = MSE(\epsilon^{k}, \mathbf{U}(\hat{\alpha_{k}}\mathbf{a}^0+\hat{\beta_{k}}\epsilon^{k},k,\mathbf{E}(\mathbf{p})\oplus\mathbf{P}(\mathbf{h}))).
\end{equation}where $\epsilon^{k}$ is the added noise and $\hat{\alpha_{k}}$, $\hat{\beta_{k}}$ are functions of $k$.

\textbf{During evaluation phase}, actions are generated by denoising random Gaussian noise into actions $\mathbf{a}$ using the trained diffusion network $\mathbf{U}$. Starting from noisy input $\mathbf{a}^{K}$, the U-Net $\mathbf{U}$ refines the noise over $K$ iterations to produce the final noise-free action $\mathbf{a}^0$:
\begin{equation}
\mathbf{a}^{k-1}=\alpha_{k}(\mathbf{a}^{k}-\gamma_{k}\mathbf{U}(\mathbf{a}^{k}, k, \mathbf{E}(\mathbf{p})\oplus \mathbf{P}(\mathbf{h})+\sigma_k\mathcal{N}(0,\mathbf{I})),
\end{equation}
where $\mathcal{N}(0, \mathbf{I})$ is Gaussian noise, and $\alpha_k$, $\gamma_k$, and $\sigma_k$ are functions of $k$, determined by the noise scheduler \cite{Ze2024DP3, chi2023diffusionpolicy}.

\textbf{Spatially Equivariant Training Strategy.}
To prevent overfitting to specific trajectories, especially with limited data, we introduce a spatially equivariant training strategy to help the model learn the spatial relationships between the gripper and manipulated objects.

\begin{figure}[t]
	\centering
	\includegraphics[trim = 15mm 162mm 15mm 59mm, clip, width=0.5\textwidth, height=110pt]{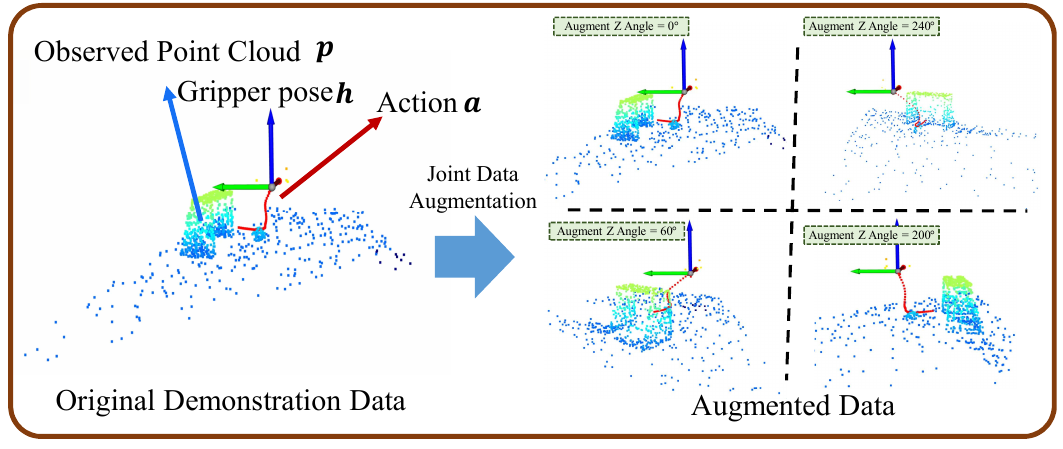}
	\caption{Given the original point cloud $\mathbf{p}$, gripper pose $\mathbf{h}$, action $\mathbf{a}$, we apply a random rotation $\mathbf{T}$ to get the augmented data $\mathbf{T}(\mathbf{p})$, $\mathbf{T}(\mathbf{h})$, and corresponding $\mathbf{T}(\mathbf{a}).$}
	\vspace{-20pt}
	\label{fig:augument}
\end{figure}

As shown in Fig. \ref{fig:augument}, after generating the point cloud $\mathbf{p}$ from SGP, we apply a random rotation $\mathbf{T}$ to the point cloud, gripper pose $\mathbf{h}$, and corresponding actions $\mathbf{a}$. This augmentation ensures that the relative spatial relationship between the gripper and the objects remains consistent, which is crucial for generalizing manipulation patterns. In essence, when the manipulated objects undergo rotation, the corresponding actions must also rotate in the same manner to ensure successful manipulation. By focusing on relative positioning rather than absolute poses, the model learns to adapt to variations in object orientation and positioning.

This approach significantly increases the training data while preserving spatial consistency. Given the augmented inputs $\mathbf{T}(\mathbf{p})$, $\mathbf{T}(\mathbf{h})$, and $\mathbf{T}(\mathbf{a})$, the training objective becomes:
\vspace{-3pt}
\begin{equation}
    \mathcal{L} = MSE(\epsilon^{k}, \mathbf{U}(\hat{\alpha_{k}}\mathbf{T}(\mathbf{a}^0)+\hat{\beta_{k}}\epsilon^{k},k,\mathbf{E}(\mathbf{T}(\mathbf{p}))\oplus\mathbf{P}(\mathbf{T}(\mathbf{h})))).
\vspace{-5pt}
\end{equation}




\subsection{Implementation}
In this section, we detail the implementation of our framework. For the SGP module, we use SAM \cite{kirillov2023segment} to segment task-relevant objects in the first frame, after which the object mask is automatically tracked using a video segmentation model \cite{cheng2024putting}. For point sampling, we sample 1,024 points from the entire scene, with an equal number of points sampled from each category.

For the SGD module, the encoder $\mathbf{E}$ follows the same point cloud extractor structure as in \cite{Ze2024DP3}, and the encoder $\mathbf{P}$ is an MLP network with channel sizes of 9, 64, and 64. During training, the diffusion model is set with a training horizon of 4, an observation step of 2, and an action step of 4. Each manipulation task is trained for 3,000 epochs.

\begin{table*}[t]
\caption{Performance comparison of our framework against other methods including DP3 \cite{Ze2024DP3}, ManiCM\cite{lu2024manicm} , Diffusion Policy\cite{chi2023diffusionpolicy}.}
\vspace{-4pt}
	\centering
	\scalebox{0.95}{
		\begin{tabular}{lcccccc}
			\toprule
			Alg. \ Task & $Dial~Turn\left(\uparrow \right)$ &  $Disassemble\left(\uparrow \right)$ & $Coffee~Pull\left(\uparrow \right)$ & $Soccer\left(\uparrow \right)$ & $Sweep~Into\left(\uparrow \right)$& $Hand~Insert\left(\uparrow \right)$\\
			\midrule
			DP3\cite{Ze2024DP3} & 66.5 & 69.7 &87.4 & 18.7 & 38.1 & 25.6\\
			ManiCM\cite{lu2024manicm} & 85.3 &70.6 &93.4 & 22.6 & 40.3 & 25.4\\
			Diffusion Policy\cite{chi2023diffusionpolicy} & 63.2 & 43.7 &34.1 & 14.9 & 10.6 & 9.7\\
			Ours & \textbf{100.0} &\textbf{100.0} &\textbf{100.0}& \textbf{92.4} & \textbf{100.0} &\textbf{100.0} \\
            \midrule
            Task & $Pick~Place\left(\uparrow \right)$ &  $Push\left(\uparrow \right)$ & $Shelf~Place\left(\uparrow \right)$ & $Stick~Pull\left(\uparrow \right)$ & $Pick~Place~Wall\left(\uparrow \right)$& $Avg.\left(\uparrow \right)$\\
			\midrule

            DP3\cite{Ze2024DP3} & 56.1 & 51.3 &48.6 &27.6 & 45.6 & 48.7\\
			ManiCM\cite{lu2024manicm} & 55.6 &55.1 &55.9 & 64.0 & 46.9 & 55.9 \\
			Diffusion Policy\cite{chi2023diffusionpolicy} & 0.0& 30.9 &11.4 & 11.2 & 5.1 &21.35\\
			Ours & \textbf{100.0} &\textbf{95.6} &\textbf{95.4}& \textbf{89.8} & \textbf{100.0}&\textbf{97.56}\\
   
			
			\bottomrule
		\end{tabular}
  
	}
 \vspace{-10pt}
	\label{tab:simulation_res}
\end{table*}
\vspace{-7pt}

\begin{table}[t]
\caption{We evaluated the performance of our framework on different ablation study variants. Here, we use the average success rate of all tasks as metric to evaluate the framework performance under different setting.}
	\centering
	\scalebox{1.5}{
		\begin{tabular}{lc}
			\toprule
			Method & $Avg.~ Success ~Rate\left(\uparrow \right)$ \\
			\midrule
			Ours-wo/cat & 61.4 \\
                Ours-wo/ref & 43.5 \\
                Ours-wo/aug & 43.2 \\
                Ours        & 97.6        \\
			
			\bottomrule
		\end{tabular}

	}
  \vspace{3pt}
 \vspace{-20pt}
	
	\label{tab:ablation}
\end{table}

\section{Experiments}

In this section, we design comprehensive simulation and real-world experiments to validate the effectiveness and efficiency of our framework. Our results demonstrate that: 1) Our framework achieves impressive performance with only 10 demonstrations across diverse manipulation tasks, outperforming current state-of-the-art methods. 2) It generalizes well to different 3D initialization challenges across a variety of tasks. 3) Our framework can be successfully applied to real-world hardware systems.

\subsection{Experiment Setup}
For the simulation experiments, we use MetaWorld \cite{yu2020meta}, a widely adopted manipulation benchmark with diverse tasks involving interactions with various objects. We select the 10 most challenging tasks, which have a success rate of less than 50\% in state-of-the-art methods \cite{Ze2024DP3,lu2024manicm,chi2023diffusionpolicy}, as the primary benchmark for evaluating our framework. For each task, we train our framework using 10 demonstrations and report the corresponding success rates.
For real-world experiments, we select five tasks to further demonstrate the effectiveness of our framework in real-world applications.

\textbf{Baselines}
This work focuses on learning generalized manipulation skills using 3D representations with few-shot demonstrations. Accordingly, our main baselines are the state-of-the-art point cloud-based 3D diffusion policies DP3 \cite{Ze2024DP3} and ManiCM\cite{lu2024manicm}, both of which emphasize learning manipulation skills from limited demonstrations:
\begin{itemize}
    \item DP3 \cite{Ze2024DP3} uses point clouds as input, sampling sparse points from the scene and compacting them into a visual feature. It then directly trains the diffusion model using the trajectories.
    \item ManiCM \cite{lu2024manicm} builds on the DP3 framework, focusing on reducing inference time to generate smoother manipulation skills. It also shows an improved success rate compared to DP3.
    \item Diffusion Policy \cite{chi2023diffusionpolicy}, the original diffusion model applied to robotics, is also included for comparison.

\end{itemize}

\subsection{Overall Performance}
We conducted 60 trials for each task, using different random seeds (0, 1, 2), and calculated the success rate for each task. The performance of our framework, along with the baselines, is shown in Table 1. The results demonstrate that our framework consistently achieves a success rate of over 80\% across all tasks. Compared to the baselines, our method shows more than a 60\% improvement in most tasks.

Examining the performance of previous state-of-the-art (SOTA) methods, we observe particularly low success rates in tasks such as Soccer, Hand Insert, Pick Place, and Stick Pull, where success rates are often below 30\%. This is likely because these tasks rely heavily on the correct trajectory, which depends on the initial and target poses of the objects—conditions that differ from the training data. With limited data, these SOTA methods fail to capture the spatial relationships between objects, leading to overfitting to specific trajectories and poor performance in testing.

In contrast, our method effectively learns the spatial knowledge embedded in the task and generates trajectories based on the observed object poses. As a result, it performs remarkably well, even in these challenging tasks.

\subsection{Ablation study:} 
To assess the contributions of each module within our framework, we conducted a series of ablation experiments. Our framework has three key components that require thorough evaluation: 1) the selection of the point cloud's coordinate reference, 2) category-based sampling of the point cloud, and 3) data augmentation during the training phase:

\begin{itemize}
    \item Ours-wo/cat: In this setting, we abandon equal sampling for each category of task-relevant objects. Instead, we sample points uniformly from the entire scene for visual embedding.
    \item Ours-wo/ref: Here, we do not use the gripper base as the reference frame. Instead, we use the fixed robot base as the reference frame to align the point cloud.
    \item Ours-wo/aug: In this variant, we exclude the data augmentation strategy and train the model using only the original data from the demonstration set.

\end{itemize}

The results of the ablation study are shown in Table \ref{tab:ablation}. The performance of each variant indicates that removing any component leads to a noticeable decline in success rates, underscoring the importance and effectiveness of each module in our framework.

\begin{table}[h]
	\caption{We continually expand the random initial region size for manipulated objects pose initialization. The success rate of each tasks are reported.}
 \vspace{-5pt}
	\centering
	\scalebox{1.3}{
		\begin{tabular}{lcccc}
			\toprule
			Push & $1$ & $2$& $4$ & $6$ \\
			\midrule
			DP3\cite{Ze2024DP3} & 51.3 & 30.0 & 20.0 & 25.0 \\

                Ours        & 98.6 & 85 & 80 & 85.7       \\
			
			\bottomrule
   		Coffee~Pull & $1$ & $2$& $4$ & $6$ \\
			\midrule
			DP3\cite{Ze2024DP3} &  87.4 & 30.0& 20.0  & 10.0  \\

                Ours        & 100 & 100 & 100 & 68.18      \\
			
			\bottomrule

                		Pick~Place & $1$ & $2$& $4$ & $6$ \\
			\midrule
			DP3\cite{Ze2024DP3} 
   & 56.1 & 0 & 0 & 0
   \\

                Ours        &96.0 &56.0 &50.0 & 44.4       \\
			
			\bottomrule
		\end{tabular}

	}
  \vspace{3pt}
 \vspace{-20pt}
	
	\label{tab:region}
\end{table}

\subsection{Generalization Ability Evaluation}
In this section, we demonstrate the generalization ability of our framework, which, despite being trained on limited data, can effectively handle challenges posed by varying initial and target poses of manipulated objects. We use three challenging tasks $Push$, $Coffee~Pull$, and $Pick~Place$ to showcase this capability.

To assess the robustness of the methods, we progressively expand size of the initialization region for each task and observe the performance under different region sizes. The initialization region refers to the area within which the manipulated object is randomly placed. All tasks were initially trained with data from the original region size, after which we expanded the region to 2x, 4x, and 6x the original size in evaluation. As the region size increases, the robot encounters a wider range of initial and target poses compared to the training data.

The results, presented in Tab.\ref{tab:region}, show that as the region size increases, our method experiences a slight performance drop. In contrast, the baseline DP3 exhibits a significant performance decline, highlighting that our approach is more robust in handling spatial variations.

\subsection{Generalization in Viewpoint}

Here, we demonstrate the strong generalization ability of our framework when dealing with changes in camera viewpoint. We use the tasks $Coffee~Pull$, $Hand~Insert$ and $Sweep~Into$ as representative examples to illustrate this. Specifically, we first train the model on data captured from a fixed camera viewpoint. In the testing phase, we apply both rotation and translation perturbations to the camera, then evaluate the tasks and record the success rates. The experimental results are shown in Tab.~\ref{tab:viewpoint}, where we list the success rates for each task under different viewpoint changes and compare them with the baseline method, DP3\cite{Ze2024DP3}.

The results clearly demonstrate that our framework has a strong generalization ability for both rotation and translation of the camera viewpoint. In contrast, previous methods, as reported in earlier works, tend to show limited tolerance to small viewpoint changes and struggle to generalize across significant angular differences.

\begin{figure}[ht]
	\centering
	\includegraphics[trim = 35mm 162mm 44mm 48mm, clip, width=0.5\textwidth, height=157pt]{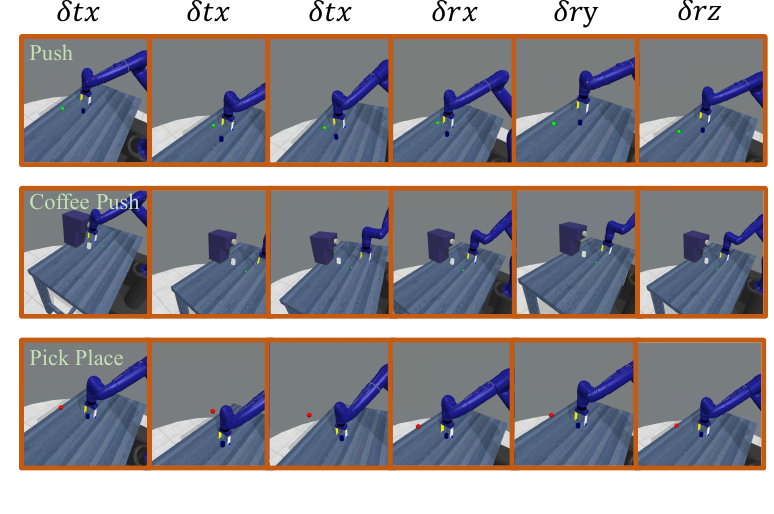}
	\vspace{-25pt}
	\caption{Here we visualize the different camera viewpoint settings used in evaluation. We apply movement along each of the six degrees of freedom, including three translations and three rotations, resulting in a total of six distinct settings.}
	\vspace{-10pt}
	\label{fig:generalization}
\end{figure}
\begin{table}[h]
	\caption{Performance of our framework on different camera viewpoints. Here, we show the success rate of each task under different setting.}
 \vspace{-4pt}

	\centering
	\scalebox{1.1}{
		\begin{tabular}{lcccccc}
			\toprule
			Push & $\delta tx$ & $\delta ty$& $\delta tz$ & $\delta rx$ & $\delta ry$ & $\delta rz$\\
			\midrule
			DP3\cite{Ze2024DP3} & 20 & 0 & 0 & 0 & 10 & 65 \\

                Ours        & 100 & 100 & 100 & 100 & 100 & 100        \\
			
			\bottomrule
   		Coffee~Pull & $\delta tx$ & $\delta ty$& $\delta tz$ & $\delta rx$ & $\delta ry$ & $\delta rz$\\
			\midrule
			DP3\cite{Ze2024DP3} &  30 & 0& 0  & 20 &90&70 \\

                Ours        & 100 & 100 & 100 & 100 & 100 & 100       \\
			
			\bottomrule

                		Pick~Place & $\delta tx$ & $\delta ty$& $\delta tz$ & $\delta rx$ & $\delta ry$ & $\delta rz$\\
			\midrule
			DP3\cite{Ze2024DP3} 
   & 0 & 0 & 0 & 0 & 0 & 0
   \\

                Ours        &80 &100 &95 & 100 & 90 & 75       \\
			
			\bottomrule
		\end{tabular}

	}
  \vspace{3pt}
 \vspace{-20pt}
	
	\label{tab:viewpoint}
\end{table}

\subsection{Real World Experiments}

We conducted several representative tasks on a real-world system to demonstrate the effectiveness of our method beyond simulation.
Our setup includes a UR5e robotic arm, a parallel gripper as the end-effector, and an RGB-D camera to capture the input images. We successfully executed the task such as $Pick~Place, Push, Sweep~Into$. For detailed results and videos of our real-world experiments, please refer to our supplementary materials and our webpage\footnote{https://github.com/renyu2016/Generalized-3D-Manipulation}. 



\section{Conclusion}
In this paper, we propose a framework that could learn generalizable 3D manipulation within 10 demonstrations. Unlike traditional methods that often overfit to limited trajectories, our framework captures the underlying spatial relationships between the gripper and manipulated objects, enabling it to handle a wide range of spatial variations. This is achieved through our spatially equivariant training strategy, where we augment manipulation trajectories in 3D space, similar to data augmentation techniques in object recognition. Through comprehensive experiments, we demonstrated both the effectiveness and efficiency of our proposed framework, showing its strong generalization capabilities and performance improvements across various challenging tasks. To the best of our knowledge, this is the first work to apply trajectory augmentation in 3D manipulation learning, and we believe this technique is as impactful and effective for manipulation tasks as rotation augmentation is for object recognition.

\vspace{-7pt}
\section*{ACKNOWLEDGMENT}
\vspace{-1pt}
This work was supported by the Major Project of the National Key Research and Development Program of China under Grant (2023YFB4704800). This work was conducted during a visiting period at College of Automation Science and Engineering, South China University of Technology.



	\bibliographystyle{IEEEtran}
	\bibliography{IEEEexample}

\end{document}